\pdfoutput=1

\documentclass[11pt]{article}

\usepackage[]{acl}

\usepackage{times}
\usepackage{latexsym}

\usepackage[T1]{fontenc}

\usepackage[utf8]{inputenc}

\usepackage{microtype}

\usepackage{inconsolata}

\usepackage{multirow}
\usepackage{booktabs}
\usepackage{graphicx}
\usepackage{subfigure}

\newcommand{\Model}{EcomGPT-CT}

\title{\Model{}: Continual Pre-training of E-commerce \\ Large Language Models with Semi-structured Data}

\author{Shirong Ma$^{1}$\thanks{ $^*$ Equal first authorship. This work was conducted during Shirong's internship at Alibaba Group.},~Shen Huang$^{2*}$,~Shulin Huang$^{1*}$,~Xiaobin Wang$^{2}$,~Yangning Li$^{1}$,\\ ~\textbf{Hai-Tao Zheng}$^{1, 3}$\thanks{ $^{\dagger}$ Corresponding authors: Yong Jiang and Hai-Tao Zheng. (E-mail: jiangyong.ml@gmail.com, zheng.haitao@sz.tsinghua.edu.cn)},~\textbf{Pengjun Xie}$^{2}$,~\textbf{Fei Huang}$^{2}$ \and \textbf{Yong Jiang}$^{2\dagger}$ \\
        $^{1}$Tsinghua Shenzhen International Graduate School, Tsinghua University \\ 
        $^{2}$Alibaba Group, $^{3}$Peng Cheng Laboratory}

\begin{document}
\maketitle
\begin{abstract}
Large Language Models (LLMs) pre-trained on massive corpora have exhibited remarkable performance on various NLP tasks.
However, applying these models to specific domains still poses significant challenges, such as lack of domain knowledge, limited capacity to leverage domain knowledge and inadequate adaptation to domain-specific data formats.
Considering the exorbitant cost of training LLMs from scratch and the scarcity of annotated data within particular domains, in this work, we focus on domain-specific continual pre-training of LLMs using E-commerce domain as an exemplar.
Specifically, we explore the impact of continual pre-training on LLMs employing unlabeled general and E-commercial corpora.
Furthermore, we design a mixing strategy among different data sources to better leverage E-commercial semi-structured data.
We construct multiple tasks to assess LLMs' few-shot In-context Learning ability and their zero-shot performance after instruction tuning in E-commerce domain.
Experimental results demonstrate the effectiveness of continual pre-training of E-commerce LLMs and the efficacy of our devised data mixing strategy.
\end{abstract}

\section{Introduction}
Large Language Models (LLMs) are pre-trained on billions of text tokens, allowing them to acquire extensive general knowledge and grasp the intricacies of human language rules~\cite{zhao2023survey}.
Consequently, LLMs have demonstrated exceptional performance across a wide range of Natural Language Processing (NLP) tasks~\cite{brown2020language, ouyang2022training, huang2023lateval}.

However, existing LLMs are not perfect, and applying them to specific domains still presents significant challenges:
(1) LLMs lack the necessary domain-specific knowledge or struggle to utilize relevant knowledge to address practical tasks~\cite{huang2023lawyer, wang2023huatuo}.
(2) LLMs face difficulties in adapting to the unique text formats or data distributions in specific domains, thus failing to meet the requirements of domain applications~\cite{li2023ecomgpt, li2023effectiveness}.

Due to the extensive hardware and lengthy training time required, it is practically infeasible to pre-train an LLM from scratch for a specific domain.
Moreover, annotated data within a specific domain is often scarce and costly, while unlabeled data is more abundant and easily accessible.
Therefore, in this work, we focus on E-commerce domain and investigate continual pre-training of LLMs to adapt them to a specific domain.

There is ample evidence to suggest that continual pre-training in specific domains significantly enhance the performance of pre-training Masked Language Models (MLMs, e.g. BERT~\cite{devlin-etal-2019-bert}, RoBERTa~\cite{liu2019roberta}) on downstream tasks in the corresponding domains~\cite{gururangan-etal-2020-dont, gu-etal-2020-train, ke2022continual, li-etal-2022-learning-dictionary}.
In the era of LLM, limited by data and computational resources, most efforts of LLM domain adaptation have been achieved through instruction tuning or prompt engineering~\cite{singhal2023towards, cui2023chatlaw, yunxiang2023chatdoctor, yu2023seqgpt}.
Only few studies attempt to introduce domain knowledge into models through continual pre-training~\cite{huang2023lawyer, wen2023chathome}, and there is a lack of in-depth analysis of domain-specific continual pre-training for LLMs from the perspective of practical applications.

In this paper, based on BLOOM~\cite{scao2022bloom}, a series of multilingual auto-regressive models, We construct \Model{} models in E-commerce domain. we explore the impact of continual pre-training for LLMs, employing general and E-commercial unlabeled corpora when complete original pre-training data is difficult to obtain.
Specifically, we focus on the performance of addressing practical E-commercial tasks and analyze the performance variations during the training process.
Additionally, we propose a data mixing strategy with different data sources to effectively leverage the abundant semi-structured data in e-commerce domain, thereby enhancing the scale and diversity of pre-training data.

In order to comprehensively assess the performance of LLMs in E-commerce domain, we construct two evaluation benchmarks containing various NLP tasks based on an E-commercial instruction dataset, EcomInstruct~\cite{li2023ecomgpt}.
These benchmarks are designed to evaluate the few-shot In-context Learning (ICL)~\cite{brown2020language} capability and zero-shot performance of models after instruction tuning.
Moreover, we select several general NLP tasks to evaluate the general few-shot ICL ability.

The experimental findings are as follows:
(1) Continual pre-training on unlabeled data in the e-commerce domain enhances model performance on domain-specific tasks. Furthermore, integrating general pre-training data with domain-specific data from e-commerce strengthens the model's domain adaptability without significantly compromising its performance on general NLP benchmarks.
(2) During the continual pre-training process, as opposed to sampling from different data sources independently, employing our designed data mixing strategy, which amalgamates data from various sources into a single training sample, enables the model to learn the intrinsic connections between different types of data more effectively, thereby achieving superior domain performance.
(3) The continual pre-training process yields variable performance gains across different tasks. Predominantly, tasks that are highly dependent on domain knowledge or exhibit unique data formats demonstrate greater performance discrepancies before and after continual pre-training.

Our contributions is summarized as follows:
(1) We conduct a thorough analysis of domain-specific continual pre-training for LLMs, and construct \Model{} models in E-commerce domain.
(2) We propose a data mixing strategy for leveraging semi-structured data from various data sources.
(3) We construct benchmarks for comprehensively evaluating the performance of few-shot ICL capability and zero-shot performance after instruction tuning in e-commerce domain.

\section{Related Work}
\subsection{Large Language Models}
In recent years, the field of Large Language Models (LLMs) has witnessed rapid advancements. Current mainstream LLMs are predominantly construct upon Transformer blocks and are pre-trained on massive amounts of text data.
Since GPT-2~\cite{radford2019language} and T5~\cite{raffel2020exploring} demonstrate that various NLP tasks can be unifed under the paradigm of text generation, prevailing LLMs decoder-only auto-regressive architecture.
Recently, following the scaling law~\cite{kaplan2020scaling}, various LLMs with different parameter sizes have been construct and released, including GPT-3~\cite{brown2020language}, Chinchilla~\cite{hoffmann2022training}, BLOOM~\cite{scao2022bloom}, PaLM~\cite{chowdhery2022palm}, Llama~\cite{touvron2023llama}, Baichuan~\cite{yang2023baichuan}.
Motivated by the remarkable performance of LLMs, researchers endeavor to construct domain-adapted LLMs for addressing tasks within particular domains, such as Med-PaLM~\cite{singhal2023towards} and ChatDoctor~\cite{yunxiang2023chatdoctor} for biomedicine, Minerva~\cite{lewkowycz2022solving} for mathematics, BloombergGPT~\cite{wu2023bloomberggpt} and FinGPT~\cite{yang2023fingpt} for finance, as well as ChatLaw~\cite{cui2023chatlaw} for law.

\begin{figure*}
    \centering
    \includegraphics[width=0.85\linewidth]{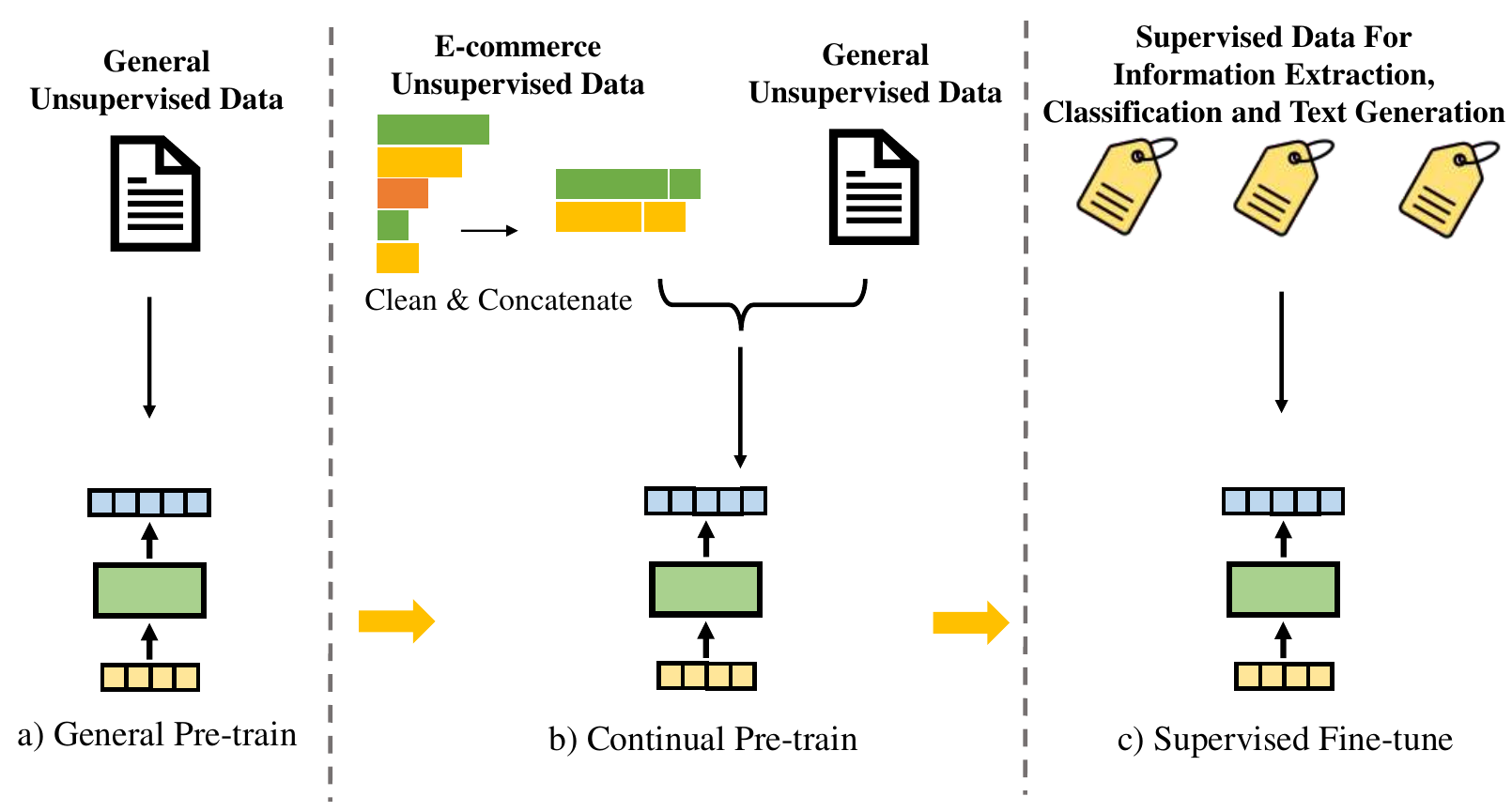}
    \caption{General framework of model training.}
    \label{fig:framework}
\end{figure*}

\subsection{Continual Pre-training}
Continual pre-training refers to an additional pre-training phase between the two stages of general pre-training and fine-tuning in downstream tasks.
Continual pre-training utilizes the same training objective as general pre-training, but it typically employs unlabeled corpora specific to a particular domain or task, aiming to achieve domain adaptation or task adaptation.
For MLM pre-training models such as BERT~\cite{devlin-etal-2019-bert} and RoBERTa~\cite{liu2019roberta}, domain-adaptive or task-adaptive continual pre-training guide the models to align with data distribution of the corresponding domain or task, thereby effectively improving the model's performance on related downstream tasks~\cite{gururangan-etal-2020-dont, gu-etal-2020-train, ke2022continual}.
Regarding current mainstream auto-regressive LLMs, there are limited studies on the process of continual pre-training for LLMs~\cite{gupta2023continual} or application of continual pre-training in constructing domain-specific LLMs~\cite{huang2023lawyer}.

However, there is a lack of detailed analysis of the impact of continual pre-training on LLMs' efficacy in addressing domain-specific tasks, and there is a dearth of research on strategies to enhance the effectiveness of LLM continual pre-training.

\section{Domain-specific Continual Pre-training}

\subsection{Training Task}
The continual pre-training seamlessly connects the stages of general pre-training and supervised fine-tuning, enhancing the model performance in specific domains or tasks, as shown in Figure~\ref{fig:framework}. Continual pre-training amplifies the training process with the same objective as general pre-training while focusing on unlabeled corpora tailored to specific domains or tasks.
In this paper, we utilize decoder-only transformer models, which represents the current mainstream LLMs architecture. Therefore, the pre-training objective is the next token prediction task or the auto-regressive language modeling:
\begin{equation}
    \mathop{\max}\limits_{\theta}{\sum_{i=1}^{N}{\log{P(y_i|y_{<i};\theta)}}},
\end{equation}
where $\theta$ is the model parameters and $y$ is a training text sequence.

\subsection{Dataset Construction}
In this paper, we focus on enhancing the performance of LLMs in the e-commerce domain through continual pre-training. We curate a large amount of e-commerce data for injecting domain-specific knowledge into LLMs and facilitating LLMs' adaptation to the distinctive text format in the e-commerce field.
Furthermore, in order to preserve the world knowledge and generalization ability of LLMs, we also incorporate general text data into the training datasets.

\paragraph{E-commerce Corpora}

We collect titles, properties, descriptions and reviews of a massive amount of products from Amazon\footnote{https://www.amazon.com/} and Taobao\footnote{https://www.taobao.com/}. In addition, articles from the Guangguang channel of Taobao are utilized that introduce product features, share user experiences, and showcase brand stories.

\paragraph{General Corpora}

To construct the general corpora, we sample texts from WuDaoCorpora~\cite{Yuan2021WuDaoCorporaAS} and RefinedWeb~\cite{Penedo2023TheRD}, which consist of 72 billion Chinese characters and 600 billion English tokens collected from web pages, respectively.

\paragraph{Data Pre-processing}
To ensure the quality of our pre-training data, we have implemented a comprehensive pipeline that incorporates data reformulation, filtering and deduplication processes. For semi-structured product data, we first group the titles, properties, descriptions, and reviews of certain products by product id.

\begin{table}
\small
\centering
\scalebox{0.92}{
\begin{tabular}{lll} 
\toprule
Dataset & Tokens & Feature \\ 
\midrule
Product Information & 10.1B & Very Short Text, Key-Value \\
Product Review & 4.0B & Short Text \\
Guangguang & 10.4B & Mid-Length Text \\ 
\midrule
WudaoCorpora & 34B & Chinese, Long Text, Web Page \\
RefinedWeb & 513B & English, Long Text, Web Page \\
\bottomrule
\end{tabular}
}
\caption{E-commerce and general datasets for continual pre-training.}
\label{tab:data}
\end{table}

The summary information of processed e-commerce and general datasets are presented in Table~\ref{tab:data}.
We observe a significant discrepancy in the number of tokens among different data sources in our collected dataset. The filtered and processed e-commerce data contains approximately 20B tokens, accounting for less than 5\% of the tokens in general domain web data. Moreover, within the general data, the quantity of Chinese data is far less than that of English data.
Considering our goal of domain-specific continual pre-training is to enhance the model's performance in addressing domain-specific tasks without compromising its generalization capabilities, it is imperative to maintain a balance in the token count between general and domain-specific data in the training samples. 
However, due to hardware limitations, we are unable to conduct detailed experiments on the proportions of different types of data. Following initial explorations, we set a 2:1 ratio for the number of tokens between general and domain-specific data, and a 1:1 ratio between Chinese and English data within the general data.

\subsection{Dataset Mixing Strategy}

Generally, commonly used data for LLM pre-training is obtained from organized long text on web pages, such as CommonCrawl and Wikipedia.
However, in some specific domains including e-commerce domain, a substantial amount of text data is stored in a semi-structured format within tables or databases.
These semi-structured text data significantly differ in form from regular text data.
Nevertheless, we believe that appropriately incorporating the data into continual pre-training can further enhance the domain-specific performance of LLMs.
To effectively transform the abundant semi-structured data into text sequences for model training, we devise a dataset mixing strategy across various data sources.

Specifically, our data mixing strategy consists of the following steps:

(1) Collect semi-structured data from various data sources, where each data source represents a set of nodes, with each node corresponding to a single data entry (i.e., a row or an object).

(2) Establish edges between nodes that are associated across the two data sources, constructing a heterogeneous graph that represents the data relationships. In our experiments, all the data is related to e-commerce products, and we utilize the product ID as a unique identifier to connect all nodes associated with the same product.

(3) Iteratively select connected clusters from the graph, adhering to the rule of covering as many data sources as possible, based on a pre-defined cluster size range. After selecting a cluster, remove all corresponding nodes from the graph to avoid redundancy.

(4) Randomly permute all nodes within each selected cluster. Extract available text from each node and concatenate them to build a training sample.

\begin{figure}
    \centering
    \includegraphics[width=0.95\linewidth]{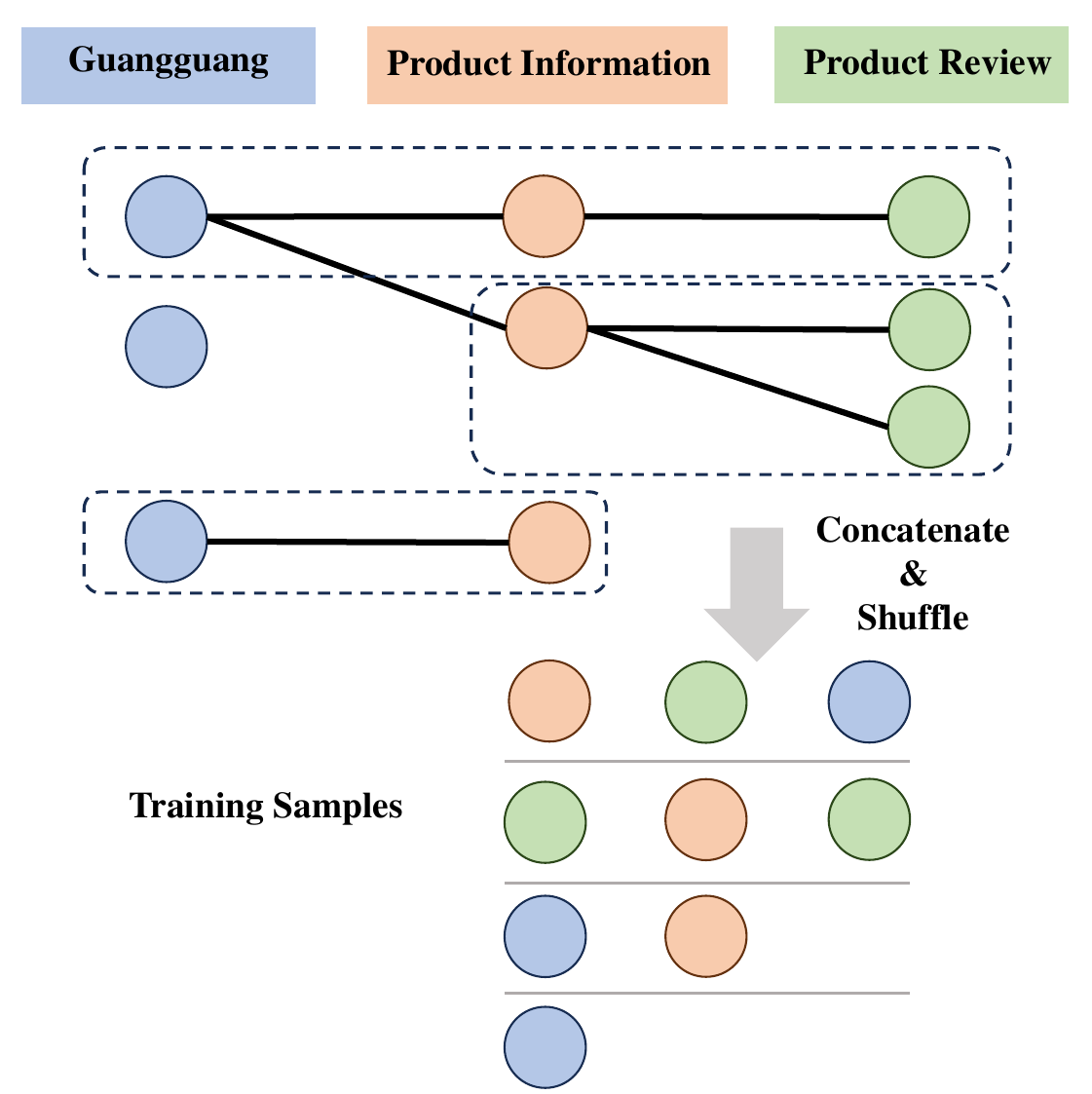}
    \caption{An example of our dataset mixing strategy.}
    \label{fig:mixing_example}
\end{figure}

An example is illustrated in Figure~\ref{fig:mixing_example}. 
We posit that this data mixing strategy establishes connections between texts from diverse data sources. As opposed to strategies that independently sample texts from each data source, our strategy enhances the diversity of texts within a single sample, thus facilitating more effective training of LLMs.

\begin{table}
\small
\centering
\scalebox{0.9}{%
\begin{tabular}{l|cc} 
\toprule
 & \textbf{\Model{}-3B} & \textbf{\Model{}-7B} \\ 
\midrule
Backbone & BLOOM-3B & BLOOM-7B \\
Parameters & 3003M & 7069M \\
Global Batch Size & \multicolumn{2}{c}{2048} \\
Sequence Length & \multicolumn{2}{c}{2048} \\
Learning Rate & 3e-5 & 2e-5 \\
Adam Betas & \multicolumn{2}{c}{(0.9, 0.95)} \\
Weight Decay & \multicolumn{2}{c}{0.1} \\
\bottomrule
\end{tabular}
}
\caption{Hyper-parameters in our experiments.}
\label{tab:hyper}
\end{table}

\begin{table*}
\small
\centering
\begin{tabular}{llll} 
\toprule
\textbf{Benchmark} & \textbf{Original Source} & \textbf{Type} & \textbf{Task (Abbreviation)} \\
\midrule
\multirow{9}{*}{\textbf{EcomICL}} & JAVE~\cite{zhu-etal-2020-multimodal} & IE & Attribute-Value Extraction (AVE) \\
 & E-commercial NER~\cite{jie-etal-2019-better} & IE & Named Entity Detection (NED) \\
 & ELM & CLS & Entity Typing (ENT) \\
 & \multirow{3}{*}{OpenBG~\cite{deng2023construction}} & CLS & Product Classification (PDC) \\
 &  & CLS & Title-Attribute Matching (TAM) \\
 &  & GEN & Title Generation (TIG) \\
 & KOBE~\cite{chen2019towards} & GEN & Description Generation (DEG) \\
 & CEPSUM~\cite{li2020aspect} & GEN & Description Summarization (DES) \\
 & LESD4EC~\cite{gong2019automatic} & GEN & Short Title Generation (STG) \\ 
\midrule
\multirow{8}{*}{\textbf{EcomSFT}} & JAVE~\cite{zhu-etal-2020-multimodal} & IE & Attribute-Value Extraction (AVE) \\
 & \multirow{3}{*}{OpenBG~\cite{deng2023construction}} & CLS & Product Classification (PDC) \\
 &  & CLS & Title-Attribute Matching (TAM) \\
 &  & GEN & Title Generation (TIG) \\
 & E-commercial NER~\cite{jie-etal-2019-better} & IE & Named Entity Recognition (NER) \\
 & JDDC~\cite{chen2020jddc} & CLS & Intent Understanding (ITU) \\
 & CCKS2021 & CLS & Address Matching (ADM) \\
 & KOBE~\cite{chen2019towards} & GEN & Description Generation (DEG) \\ 
\midrule
\multirow{6}{*}{\textbf{GeneralICL}} & AGNews~\cite{zhang2015character} & CLS & Topic Classification \\
 & SQUAD v2~\cite{rajpurkar-etal-2016-squad} & MRC & Reading Comprehension \\
 & SNLI~\cite{bowman-etal-2015-large} & CLS & Natural Language Inference \\
 & TNews~\cite{xu-etal-2020-clue} & CLS & Topic Classification \\
 & CMRC2018~\cite{cui-etal-2019-span} & MRC & Reading Comprehension \\
 & OCNLI~\cite{hu-etal-2020-ocnli} & CLS & Natural Language Inference \\
\bottomrule
\end{tabular}
\caption{Tasks of benchmarks.}
\label{tab:benchmark}
\end{table*}

\subsection{Training Setup}
We select BLOOM~\cite{scao2022bloom} as the backbone models of \Model{} due to the following reasons: 
(1) BLOOM is a series of multilingual pre-trained models supporting both English and Chinese. (2) BLOOM has not undergone any post training, ensuring the reliability of evaluating the performance changes on our benchmarks after continual pre-training.
Specifically, We conduct our experiments on BLOOM with 3B and 7.1B parameters, respectively.

Limited by the hardware quotas, each of our experiments is conducted on 2-4 NVIDIA Tesla A100 80GB GPUs. This serves as a valuable reference for application scenarios with limited computational resources.
We employ Huggingface's Transformers~\cite{wolf2019huggingface} and DeepSpeed~\cite{rasley2020deepspeed} frameworks. Transformers provides the model implementation and fundamental training workflow, while DeepSpeed provides ZeRO optimizer~\cite{rajbhandari2020zero} for sharding the training states (e.g. model parameters, gradients, optimizer states), thereby optimizing GPU memory consumption. Specifically, we use ZeRO stage 2 with offload, which represents that the optimizer states and gradients are sharded and the training states can be swapped between the host and devices.

The training process is performed in bfloat16 mixed precision~\cite{kalamkar2019study} to enhance training efficiency and avoid numerical underflow or overflow problems.
Hyper-parameters utilized in the experiments are presented in Table~\ref{tab:hyper}. We set some of the hyper-parameters based on the original pre-training settings of BLOOM.

\section{Experiments}

\begin{table*}
\small
\centering
\begin{tabular}{l|cc|ccc|cccc} 
\toprule
 & \multicolumn{2}{c|}{\textbf{IE}} & \multicolumn{3}{c|}{\textbf{CLS}} & \multicolumn{4}{c}{\textbf{GEN}} \\
 & \textbf{AVE} & \textbf{NED} & \textbf{ENT} & \textbf{PDC} & \textbf{TAM} & \textbf{TIG} & \textbf{DEG} & \textbf{DES} & \textbf{STG} \\
 & \textbf{Acc} & \textbf{Acc} & \textbf{Acc} & \textbf{Acc} & \textbf{Acc} & \textbf{ROUGE} & \textbf{ROUGE} & \textbf{ROUGE} & \textbf{ROUGE} \\ 
\midrule
BLOOM-3B & 31.4 & 36.4 & 73.1 & 21.8 & \textbf{51.3} & 19.3 & 21.3 & 14.9 & 48.6 \\
\quad+ General & 10.0 & 34.0 & 67.1 & 7.2 & 50.4 & 19.6 & \textbf{22.1} & 14.9 & 49.7 \\
\quad+ General + Ecom (Separate) & 30.1 & 36.9 & \textbf{73.9} & 28.6 & 49.2 & 24.9 & 22.0 & 17.6 & 52.8 \\
\Model{}-3B & \textbf{32.3} & \textbf{37.3} & 73.5 & \textbf{28.9} & 49.7 & \textbf{28.7} & 21.8 & \textbf{18.1} & \textbf{53.1} \\ 
\midrule
BLOOM-7B & 33.3 & 41.6 & 71.5 & 25.0 & 51.8 & 20.8 & \textbf{22.3} & 16.2 & 53.3 \\
\quad+ General & 17.0 & 35.3 & \textbf{72.9} & 21.4 & \textbf{53.6} & 20.0 & \textbf{22.3} & 16.8 & 50.5 \\
\quad+ Ecom (Separate) & 33.1 & 41.3 & 70.4 & 28.4 & 49.7 & 25.2 & 22.0 & 18.2 & 54.5 \\
\quad+ General + Ecom (Separate) & 35.2 & \textbf{42.4} & 70.3 & 30.8 & 52.1 & 27.3 & 21.9 & 17.4 & 54.3 \\
\Model{}-7B & \textbf{37.4} & 42.1 & 70.0 & \textbf{30.9} & 52.2 & \textbf{30.5} & 22.2 & \textbf{18.7} & \textbf{55.0} \\
\bottomrule
\end{tabular}
\caption{Experiment results of EcomICL. We report the performance on 9 ICL benchmarks with various experiment settings. Note that \Model{} is equivalent to BLOOM + General + Ecom (Mixed).}
\label{tab:ecom_icl}
\end{table*}

\begin{table*}
\small
\centering
\begin{tabular}{l|cccc|cccc} 
\toprule
 & \multicolumn{4}{c|}{\textbf{Held-in}} & \multicolumn{4}{c}{\textbf{Held-out}} \\
 & \textbf{NER} & \textbf{DEG} & \textbf{ADM} & \textbf{ITD} & \textbf{AVE} & \textbf{TIG} & \textbf{TAM} & \textbf{PDC} \\
\midrule
BLOOM-3B & 72.3 & 15.8 & 79.2 & 52.1 & 81.2 & 24.5 & 84.4 & 45.8 \\
\quad+ General & \textbf{73.0} & 14.6 & 79.6 & 51.4 & 78.3 & 24.3 & \textbf{85.0} & 52.8 \\
\quad+ General + Ecom (Separate) & 72.7 & 17.0 & \textbf{79.9} & 53.6 & 79.6 & 29.3 & 84.1 & 59.4 \\
\Model{}-3B & 72.9 & \textbf{17.4} & \textbf{79.9} & \textbf{54.3} & \textbf{81.3} & \textbf{29.4} & 84.0 & \textbf{62.9} \\ 
\midrule
BLOOM-7B & 70.3 & 17.5 & \textbf{80.1} & 53.3 & \textbf{81.4} & 25.7 & \textbf{84.6} & 46.6 \\
\quad+ General & 70.3 & 17.6 & 79.3 & 53.6 & 80.1 & 24.6 & 84.4 & 51.0 \\
\quad+ Ecom (Separate) & 69.8 & \textbf{18.7} & 79.8 & 54.1 & 79.8 & 27.7 & 84.0 & 53.1 \\
\quad+ General + Ecom (Separate) & 70.7 & 18.6 & 79.5 & 54.6 & 81.0 & 30.1 & 82.6 & 54.6 \\
\Model{}-7B & \textbf{70.9} & 18.1 & 79.7 & \textbf{55.3} & 81.3 & \textbf{31.0} & 84.1 & \textbf{60.6} \\
\bottomrule
\end{tabular}
\caption{Experiment results of EcomSFT. We report the \textbf{ROUGE} metrics on 4 held-in and 4 held-out datasets.}
\label{tab:ecom_sft}
\end{table*}

\subsection{Evaluation Benchmarks}

In order to evaluate the performance of foundation LLMs in e-commerce domain from the perspective of solving practical problems, we construct two benchmarks based on EcomInstruct~\cite{li2023ecomgpt}, as presented in Table~\ref{tab:benchmark}:

\noindent(1) \textbf{EcomICL}. We select 9 tasks covering various types including text classification (CLS), text generation (GEN), and information extraction (IE).
Each data instance is processed into a standardized format and multiple examples with the same format are provided as demonstrations, to assess LLMs' few-shot In-context Learning (ICL) performance.

\noindent(2) \textbf{EcomSFT}. We perform instruction-tuning on foundation LLMs employing the training data of EcomInstruct. Eight tasks are chosen for assessing instruction-following LLMs, including 4 held-in and 4 held-out tasks, with the aim of evaluating the instruction-following performance of models after supervised fine-tuning (SFT).

Furthermore, to assess whether foundation LLMs retain the ability to address general NLP tasks after continual pre-training, we select 3 tasks from Chinese and English NLP benchmarks, respectively, and formulate them in the form of Few-shot ICL (\textbf{GeneralICL}).
For each of these tasks, up to 1000 data instances are randomly chosen in our experiments.

For auto-regressive LLMs, every task is regarded as a text generation task. Therefore, we can assess the model's performance on diverse tasks employing automatic evaluation metrics for text generation. Following previous work~\cite{muennighoff2022crosslingual, mishra-etal-2022-cross}, ROUGE-L~\cite{lin-2004-rouge} is utilized as an evaluation metric.
In addition, we report the accuracy metric for classification tasks, while we also employ precision, recall and F1 metrics for the two IE tasks.

\subsection{Results on Domain-specific Tasks}

\begin{table*}
\small
\centering
\begin{tabular}{l|ccc|ccc} 
\toprule
 & \multicolumn{3}{c|}{\textbf{English }} & \multicolumn{3}{c}{\textbf{Chinese}} \\
 & \textbf{AGNews} & \textbf{SQUAD v2} & \textbf{SNLI} & \textbf{TNews} & \textbf{CMRC2018} & \textbf{OCNLI} \\
 & \textbf{Accuracy} & \textbf{ROUGE} & \textbf{Accuracy} & \textbf{Accuracy} & \textbf{ROUGE} & \textbf{Accuracy} \\ 
\midrule
BLOOM-3B & 38.5 & 41.8 & 33.9 & 36.8 & 49.6 & 35.0 \\
\quad+ General & 39.0 & 43.8 & 34.5 & 39.1 & 48.6 & 36.0 \\
\quad+ General + Ecom (Separate) & 37.1 & 48.1 & 33.0 & 42.6 & 55.3 & 33.4 \\
\Model{}-3B & 38.3 & 45.3 & 34.4 & 40.3 & 52.1 & 35.7 \\ 
\midrule
BLOOM-7B & 38.2 & 53.5 & 33.1 & 42.6 & 59.2 & 32.8 \\
\quad+ General & 38.8 & 53.3 & 35.2 & 37.6 & 56.8 & 33.8 \\
\quad+ Ecom (Separate) & 38.1 & 52.7 & 33.1 & 37.1 & 57.4 & 34.3 \\
\quad+ General + Ecom (Separate) & 40.9 & 53.1 & 34.1 & 39.6 & 58.7 & 33.1 \\
\Model{}-7B & 41.9 & 53.3 & 35.3 & 42.5 & 59.8 & 33.6 \\
\bottomrule
\end{tabular}
\caption{Results of GeneralICL. We select NLP tasks in both English and Chinese, respectively.}
\label{tab:gen_icl}
\end{table*}

\begin{figure*}
    \centering
    \subfigure[EcomICL]{
        \includegraphics[width=0.78\columnwidth]{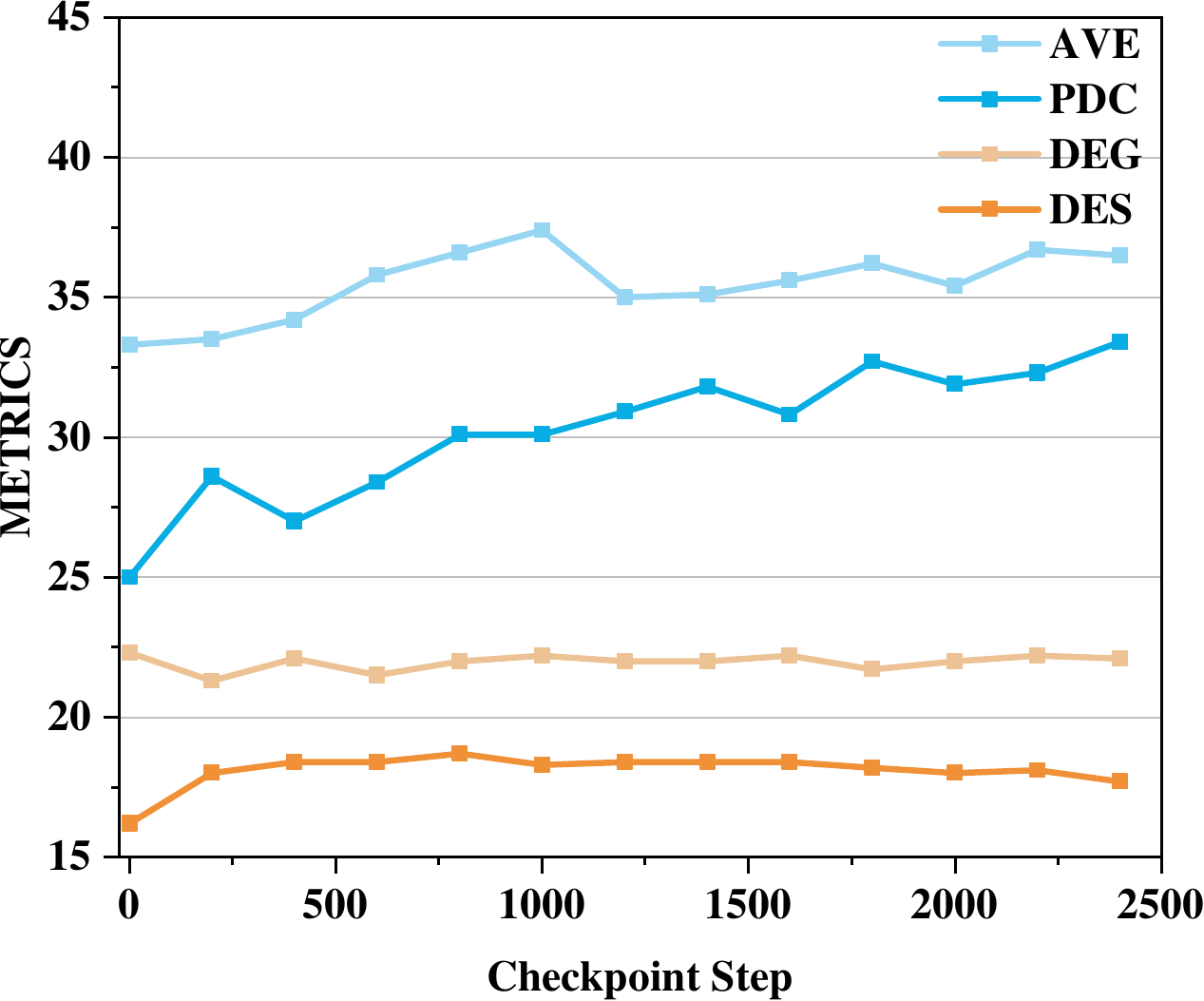}
        \label{fig:variation1}
    }
    \hspace{12mm}
    \subfigure[GeneralICL]{
        \includegraphics[width=0.78\columnwidth]{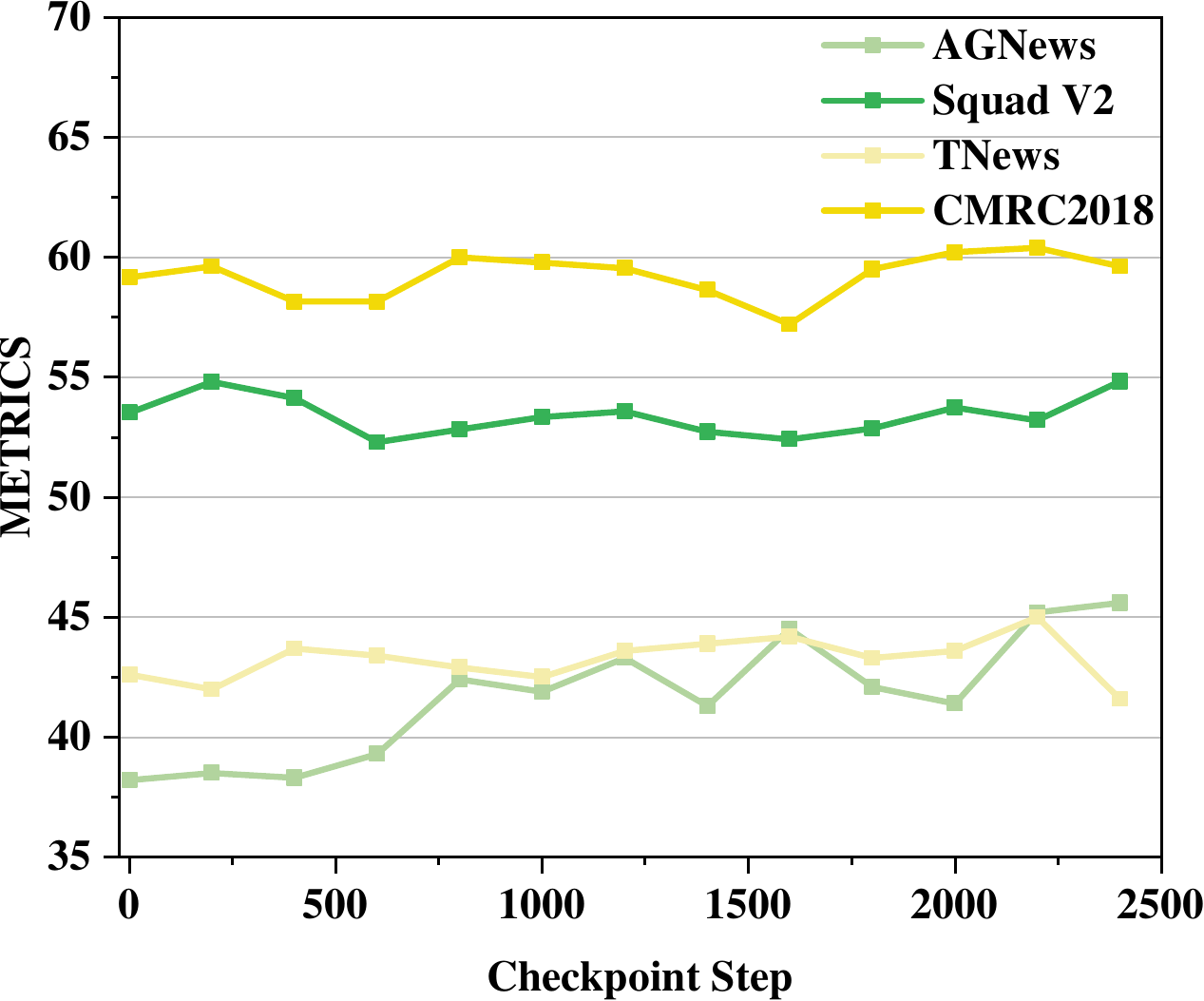}
        \label{fig:variation2}
    }
    \vspace{-3mm}
    \caption{Performance variation of \Model{}-7B during the continual pre-training process.}
\end{figure*}

Table~\ref{tab:ecom_icl} presents the evaluation results of the BLOOM-3B and 7B models on diverse E-commerce Few-shot ICL (EcomICL) tasks, considering different continual pre-training settings.
Few-shot ICL evaluation provides a direct reflection of the capabilities of LLMs in tackling domain-specific tasks without the need for additional fine-tuning.
Experiment results demonstrate that:

\noindent(1) From the perspective of training data, continual pre-training with E-commercial data effectively enhances ICL performance of 3B and 7B models on a subset of E-commerce tasks.
Meanwhile, relying solely on general data for model training results in a significant performance degradation across almost all tasks.
In addition, training with a mixture of domain-specific and general data yields more consistent performance improvements.
The results underscore the necessity of integrating both domain-specific and general corpora into continual pre-training, as domain-specific data facilitate the model's adaptation to domain-specific knowledge and data formats, while general data assist the model to avoid forgetting world knowledge and maintain generalization capabilities.

\noindent(2) Considering the aspect of data mixing, our devised strategy of integrating data from different sources into a context achieves more performance improvement or less performance degradation on most of tasks, surpassing the outcomes achieved by sampling from diverse data sources separately. Significant improvements are observed in Attribute-Value Extraction (AVE) and Title Generation (TIG) tasks.
This indicates the efficacy of our data mixing strategy in establishing interconnections among diverse data sources and infusing domain-specific knowledge from E-commercial data into the model.

\noindent(3) The influence of domain-specific continual pre-training on model performance varies across different types of tasks.
For tasks that heavily rely on domain knowledge (e.g., Product Classification (PDC) and Title Generation (TIG)) or exhibit substantial differences in data format compared to general text (e.g., Description Summary (DES) and Short Title Generation (STG)), continual pre-training significantly enhance model performance on these tasks.
Meanwhile, for tasks that require some domain knowledge but have minor disparities in form compared to general NLP tasks (e.g. Attribute-Value Extraction (AVE) and Named Entity Detection (NED)), domain-specific continual pre-training only offers marginal performance improvements.
On the contrary, for tasks that lack domain specificity (e.g. Entity Typing (ENT)) or are inherently challenging to address through Few-shot ICL (e.g. Title-Attribute Matching (TAM)), continual pre-training fails to yield effective benefits.

We also conduct supervised instruction tuning on models trained with different configurations, employing the same EcomInstruct training dataset.
Subsequently, we evaluate the performance of corresponding models after SFT across various tasks.
Instruction fine-tuned models can be utilized to address NLP problems in practical scenarios with greater convenience. Therefore, experimental results on the corresponding benchmarks indirectly reveal the impact of continual pre-training on domain-specific performance of LLMs. Table~\ref{tab:ecom_sft} presents the relevant experimental results, from which we observe the following phenomena:

\noindent(1) Similar to the results of Few-shot ICL evaluation, continual pre-training on domain-specific data effectively improves the model performance on the majority of benchmarks, and integrating general pre-training data with domain-specific data further enhances the model performance.
Moreover, during the continual pre-training process, our devised data mixing strategy similarly proves to be more effective in augmenting the model's performance after SFT, compared to separately sampling domain-specific data from diverse sources.

\noindent(2) The influence of continual pre-training on domain performance after instruction tuning also varies depending on the task type. Similar to the findings from EcomICL, continual pre-training yields substantial gains for tasks that necessitate domain knowledge or involve specific data formats (e.g., Title Generation (TIG), Product Classification (PDC), Intent Detection (ITD)), while the gains are modest for other tasks.

\noindent(3) Continual pre-training exhibits an overall performance improvement for both held-in and held-out tasks, with a more pronounced enhancement for held-out tasks. The result aligns with intuition, since SFT directly trains the model's capacity to tackle some specific tasks, thereby mitigating the impact of continual pre-training. During the process, relevant data of held-in tasks are encompassed within the training set, rendering these tasks more susceptible to the influence of SFT.

\subsection{Results on General Tasks}
In addition to domain-specific benchmarks, we conduct evaluations on Few-shot ICL performance of LLMs across several general NLP tasks in English and Chinese. Table 3 provides insights into the influence of continual pre-training on the general capabilities of LLMs.

The experimental results indicate that from the perspective of solving practical problems, continual pre-training of models with a combination of general and domain-specific data has minimal impact on the model's ability to resolve classic NLP tasks such as Topic Classification, Machine Reading Comprehension and Natural Language Inference using Few-shot ICL. 
However, since our domain-specific data are primarily in Chinese, utilizing only domain-specific data for continual pre-training results in a performance decline on Chinese benchmarks. 
This validates the critical importance of incorporating general data during domain specific pre-training to preserve the model's general NLP capabilities. 
It is worth noting that in this experiment, we do not evaluate the model's complex reasoning or knowledge retention ability. Therefore, it cannot be concluded whether domain-specific pre-training would impair these aspects of the model's capabilities.

\subsection{Performance Variation during Training}
We also assess several model checkpoints during the continual pre-training process on domain-specific and general ICL benchmarks to analyze the performance variation of LLMs during the training process.
Figure~\ref{fig:variation1} and~\ref{fig:variation2} illustrate the performance of various model checkpoints on some representative tasks.

Three distinct trends are observed from the performance changes of the four tasks in EcomICL:

(1) The model's performance steadily improves in tasks, e.g., Product Classification (PDC).
(2) The model performance remains relatively stable in some tasks, such as Description Generation (DEG).
(3) In certain tasks including Attribute-Value Extraction (AVE) and Description Summarization (DES), the model's performance initially improves but ceases to significantly change after reaching a turning point.

Regarding the four typical tasks in GeneralICL, it is evident that the model's performance fluctuates during the training process. However, within the observed time range, the metrics do not deviate significantly from the initial values before continual pre-training. The result further emphasizes that selecting appropriate data for continual pre-training effectively maintains the model's general NLP capabilities.

\section{Conclusion}
In this paper, we present an extensive investigation into the domain-specific continual pre-training of Large Language Models (LLMs), with a particular focus on the e-commerce domain. Our experiments demonstrate that LLMs can be effectively adapted to specialized domains through targeted continual pre-training using a mix of general and domain-specific corpora. Continual pre-training improves the model's performance on e-commerce tasks without sacrificing its capabilities in broader NLP applications.
Moreover, we devise a data mixing strategy, which is proven to be effective in enhancing the model's ability to assimilate and integrate knowledge from diverse data sources, resulting in improvements in domain-specific task performance.

Our work provide insights for future research in the domain adaptation of LLMs, particularly in scenarios where computational resources are constrained, and domain-specific annotated data is scarce. In the future, We plan to investigate more sophisticated data utilization strategies that yield significant performance improvements for LLMs within specific domains. Furthermore, we will continue to develop new benchmarks to assess the capabilities of unaligned foundation models in specialized domains more accurately.



\bibliography{anthology,custom}




\end{document}